\documentclass{article}

\usepackage[final,nonatbib]{neurips_2021}

\usepackage[utf8]{inputenc}
\usepackage{booktabs}

\usepackage{graphicx}
\usepackage{booktabs}
\usepackage{makecell}
\usepackage{amsmath}

\usepackage{listings}
\usepackage{color,xcolor}

\usepackage{subfig}
\usepackage[utf8]{inputenc} 
\usepackage[T1]{fontenc}    
\usepackage{url}            
\usepackage{booktabs}       
\usepackage{amsfonts}       
\usepackage{nicefrac}       
\usepackage{microtype}      
\usepackage{xcolor}         
\usepackage{multirow}
\usepackage{dcolumn}
\usepackage{xspace}
\usepackage{makecell}
\usepackage{amsmath}
\usepackage{graphicx}
\usepackage{wrapfig}
\usepackage{amssymb}
\usepackage{pifont}
\usepackage{enumitem}
\usepackage{color, colortbl}
\usepackage[ruled,vlined]{algorithm2e}
\definecolor{citecolor}{HTML}{2980b9}
\definecolor{linkcolor}{HTML}{c0392b}
\usepackage[pagebackref=true,breaklinks=true,colorlinks,bookmarks=false,citecolor=citecolor,linkcolor=linkcolor]{hyperref}

\newcommand{\model}{\mbox{\sc{Container}}\xspace}
\newcommand{\modellight}{\mbox{\sc{Container-Light}}\xspace}
\newcommand{\modelpam}{\mbox{\sc{Container-Pam}}\xspace}

\title{
Container: Context Aggregation Network}

\usepackage{mathtools}

\usepackage{macros/mysymbols}
\usepackage{macros/widetext}
\usepackage{macros/space_saver}

\author{Peng Gao$^{1,2}$
, \enskip Jiasen Lu$^{4}$, \enskip Hongsheng Li$^{2}$, \enskip Roozbeh Mottaghi$^{3,4}$, \enskip Aniruddha Kembhavi$^{3,4}$\\
$^1$ Shanghai AI Laboratory \quad  $^2$ CUHK-SenseTime Joint Lab, CUHK \quad \\ $^3$ University of Washington  $^4$ PRIOR @ Allen Institute for AI 
}

\lstdefinestyle{Python}{
    language        =   Python, 
    basicstyle      =   \small\ttfamily,
    numberstyle     =   \small\ttfamily,
    keywordstyle    =   \color{blue},
    keywordstyle    =   [2] \color{teal},
    stringstyle     =   \color{magenta},
    commentstyle    =   \color{red}\ttfamily,
    breaklines      =   true,   
    columns         =   fixed,  
    basewidth       =   0.5em,
}

\begin{document}
\maketitle

\begin{abstract}
Convolutional neural networks (CNNs) are ubiquitous in computer vision, with a myriad of effective and efficient variations. Recently, Transformers -- originally introduced in natural language processing -- have been increasingly adopted in computer vision. While early adopters continue to employ CNN backbones, the latest networks are end-to-end CNN-free Transformer solutions. A recent surprising finding shows that a simple MLP based solution without any traditional convolutional or Transformer components can produce effective visual representations. While CNNs, Transformers and MLP-Mixers may be considered as completely disparate architectures, we provide a unified view showing that they are in fact special cases of a more general method to aggregate spatial context in a neural network stack. We present the \model (CONText AggregatIon NEtwoRk), a general-purpose building block for multi-head context aggregation that can exploit long-range interactions \emph{a la} Transformers while still exploiting the inductive bias of the local convolution operation leading to faster convergence speeds, often seen in CNNs. Our \model architecture achieves 82.7 \% Top-1 accuracy on ImageNet using 22M parameters, +2.8 improvement compared with DeiT-Small, and can converge to 79.9 \% Top-1 accuracy in just 200 epochs. In contrast to Transformer-based methods that do not scale well to downstream tasks that rely on larger input image resolutions, our efficient network, named \modellight, can be employed in object detection and instance segmentation networks such as DETR, RetinaNet and Mask-RCNN to obtain an impressive detection mAP of 38.9, 43.8, 45.1 and mask mAP of 41.3, providing large improvements of 6.6, 7.3, 6.9 and 6.6 pts respectively, compared to a ResNet-50 backbone with a comparable compute and parameter size. Our method also achieves promising results on self-supervised learning compared to DeiT on the DINO framework. Code is released at \url{https://github.com/allenai/container}.

\end{abstract}
\section{Introduction}

Convolutional neural networks (CNNs) have become the \textit{de facto} standard for extracting visual representations, and have proven remarkably effective at numerous downstream tasks such as object detection~\cite{lin2017focal}, instance segmentation~\cite{he2017mask} and image captioning~\cite{Anderson2018BottomUpAT}. Similarly, in natural language processing, Transformers rule the roost~\cite{devlin2018bert,radford2015unsupervised,radford2021learning,brown2020language}. Their effectiveness at capturing short and long range information have led to state-of-the-art results across tasks such as question answering~\cite{rajpurkar2016squad} and language understanding~\cite{wang2018glue}.

In computer vision, Transformers were initially employed as long range information aggregators across space (e.g., in object detection~\cite{carion2020end}) and time (e.g., in video understanding~\cite{wang2018non}), but these methods continued to use CNNs~\cite{lecun1998gradient} to obtain raw visual representations. More recently however, CNN-free visual backbones employing Transformer modules~\cite{touvron2020training,dosovitskiy2020image} have shown impressive performance on image classification benchmarks such as  ImageNet~\cite{krizhevsky2012imagenet}. The race to dethrone CNNs has now begun to expand beyond Transformers -- a recent unexpected result shows that a multi-layer perceptron (MLP) exclusive network~\cite{tolstikhin2021mlpmixer} can be just as effective at image classification.

On the surface, CNNs~\cite{lecun1998gradient,chollet2017xception,xie2017aggregated,he2016deep}, Vision Transformers (ViTs)~\cite{dosovitskiy2020image,touvron2020training} and MLP-mixers~\cite{tolstikhin2021mlpmixer} are typically presented as disparate architectures. However, taking a step back and analyzing these methods reveals that their core designs are quite similar. Many of these methods adopt a cascade of neural network blocks. Each block typically consists of aggregation modules and fusion modules. Aggregation modules share and accumulate information across a predefined context window over the module inputs (e.g., the self attention operation in a Transformer encoder), while fusion modules combine position-wise features and produce module outputs (e.g., feed forward layers in ResNet).

In this paper, we show that the primary differences in many popular architectures result from variations in their aggregation modules. These differences can in fact be characterized as variants of an affinity matrix within the aggregator that is used to determine information propagation between a query vector and its context. For instance, in ViTs~\cite{dosovitskiy2020image,touvron2020training}, this affinity matrix is dynamically generated using key and query computations; but in the Xception architecture~\cite{chollet2017xception} (that employs depthwise convolutions), the affinity matrix is static -- the affinity weights are the same regardless of position, and they remain the same across all input images regardless of size. And finally the MLP-Mixer~\cite{tolstikhin2021mlpmixer} also uses a static affinity matrix which changes across the landscape of the input.

Along this unified view, we present \model (CONText AggregatIon NEtwoRk), a general purpose building block for multi-head context aggregation. A \model block contains both static affinity as well as dynamic affinity based aggregation, which are combined using learnable mixing coefficients.  This enables the \model block to process long range information while still exploiting the inductive bias of the local convolution operation. \model blocks are easy to implement, can easily be substituted into many present day neural architectures and lead to highly performant networks whilst also converging faster and being data efficient.

Our proposed \model architecture obtains 82.7 \% Top-1 accuracy on ImageNet using 22M parameters, improving +2.8 points over DeiT-S~\cite{touvron2020training} with a comparable number of parameters. It also converges faster, hitting DeiT-S's accuracy of 79.9 \% in just 200 epochs compared to 300.

We also propose a more efficient model, named \modellight that employs only static affinity matrices early on but uses the learnable mixture of static and dynamic affinity matrices in the latter stages of computation. In contrast to ViTs that are inefficient at processing large inputs, \modellight can scale to downstream tasks such as detection and instance segmentation that require high resolution input images. Using a \modellight backbone and 12 epochs of training, RetinaNet~\cite{lin2017focal} is able to achieve 43.8 mAP, while Mask-RCNN~\cite{he2017mask} is able to achieve 45.1 mAP on box and 41.3 mAP on instance mask prediction, improvements of +7.3, +6.9 and +6.6 respectively, compared to a ResNet-50 backbone. The more recent DETR and its variants SMCA-DETR and Deformable DETR~\cite{carion2020end,gao2021fast,zhu2020deformable} also benefit from \modellight and achieve 38.9, 43.0 and 44.2 mAP, improving significantly over their ResNet-50 backbone baselines. 

\modellight is data efficient. Our experiments show that it can obtain an ImageNet Top-1 accuracy of 61.8 using just 10\% of training data, significantly better than the 39.3 accuracy obtained by DeiT. \modellight also convergences faster and achieves better kNN accuracy (71.5) compared to DeiT (69.6) under DINO self-supervised training framework~\cite{caron2021emerging}. 

The \model unification and framework enable us to easily reproduce several past models and even extend them with just a few code and parameter changes. We extend multiple past models and show improved performance -- for instance, we produce a Hierarchical DeiT model, a multi-head MLP-Mixer and add a static affinity matrix to the DeiT architecture. Our code base and models will be released publicly. Finally, we analyse a \model model containing both static and dynamic affinities and show the emergence of convolution-like local affinities in the early layers of the network.

In summary, our contributions include: (1) A unified view of popular architectures for visual inputs -- CNN, Transformer and MLP-mixer, (2) A novel network block -- \model, which uses a mix of static and dynamic affinity matrices via learnable parameters and the corresponding architecture with strong results in image classification and (3) An efficient and effective extension -- \modellight with strong results in detection and segmentation. Importantly, we see that a number of concurrent works are aiming to fuse the CNN and Transformer architectures~\cite{Li2021LocalViTBL,Xu2021CoScaleCI,liu2021swin, Heo2021RethinkingSD, Vaswani2021ScalingLS, Zhang2021MultiScaleVL, Xu2021CoScaleCI,srinivas2021bottleneck}, validating our approach. We hope that our unified view helps place these different concurrent proposals in context and leads to a better understanding of the landscape of these methods.

\section{Related Work}
\noindent \textbf{Visual Backbones.}
Since AlexNet~\cite{krizhevsky2012imagenet} revolutionized computer vision, a host of CNN based architectures have provided further improvements in terms of accuracy including  VGG~\cite{simonyan2014very}, ResNet~\cite{he2016deep}, Inception Net~\cite{szegedy2015going}, SENet~\cite{hu2018squeeze}, ResNeXt~\cite{xie2017aggregated} and Xception~\cite{chollet2017xception} and efficiency including Mobile-net v1~\cite{howard2017mobilenets}, Mobile-net v2~\cite{howard2017mobilenets} and Efficient-net v2~\cite{Tan2021EfficientNetV2SM}.
With the success of Transformers~\cite{vaswani2017attention} in NLP such as BERT~\cite{devlin2018bert} and GPT~\cite{radford2015unsupervised}, researchers have begun to apply them towards solving the long range information aggregation problem in computer vision. ViT~\cite{dosovitskiy2020image}/DeiT~\cite{touvron2020training} are transformers that achieve better performance on ImageNet than CNN counterparts. Recently, several concurrent works explore integrating convolutions with transformers and achieve promising results. ConViT~\cite{d2021convit} explores soft convolutional inductive bias for enhancing DeiT. CeiT~\cite{yuan2021incorporating} directly incorporates CNNs into the Feedforward module of transformers to enhance the learned features. PVT~\cite{wang2021pyramid} proposes a pyramid vision transformer for efficient transfer to downstream tasks. Pure Transformer models such as ViT/DeiT however, require huge GPU memory and computation for detection~\cite{wang2021pyramid} and segmentation~\cite{zheng2020end} tasks, which need high resolution input. MLP-Mixer~\cite{tolstikhin2021mlpmixer} shows that simply performing transposed MLP followed by MLP can achieve near state-of-the-art performance. We propose \model, a new visual backbone that provides a unified view of these different architectures and performs well across several vision tasks including ones that require a high resolution input.

\noindent \textbf{Transformer Variants.}
Vanilla Transformers are unable to scale to long sequences or high-resolution images due to the quadratic computation in self-attention. Several methods have been proposed to make Transformer computations more efficient for high resolution input. Reformer~\cite{kitaev2020reformer}, Clusterform~\cite{vyas2020fast}, Adaptive Clustering Transformer~\cite{zheng2020end} and Asymmetric Clustering~\cite{daras2020smyrf} propose to use Locality Sensitivity Hashing to cluster keys or queries and reduce quadratic computation into linear computation. Lightweight convolution~\cite{wu2019pay} explore convolution architectures for replacing Transformers but only explore applications in NLP. RNN Transformer~\cite{katharopoulos2020transformers} builds a connection between RNN and Transformer and results in attention with linear computation. Linformer~\cite{wang2020linformer} changes the multiplication order of key,query,value into query,value,key by deleting the softmax normalization layer and achieve linear complexity. Performer~\cite{choromanski2020rethinking} uses Orthogonal Random Feature to approximate full rank softmax attention. MLIN~\cite{gao2019multi} 
performs interaction between latent encoded nodes, and its complexity is linear with respect to input length. 
Bigbird~\cite{beltagy2020longformer} breaks the full rank attention into local, randomly selected and global attention. Thus the computation complexity becomes linear. Longformer~\cite{zaheer2020big} uses local Transformers to tackle the problem of massive GPU memory requirements for long sequences. MLP-Mixer~\cite{tolstikhin2021mlpmixer} is a pure MLP architecture for image recognition. In the unified formulation we provide, MLP-Mixer can be considered as a single-head Transformer with static affinity matrix weight. 
MLP-Mixer can provide more efficient computation than vanilla transformer due to no need to calculate affinity matrix using key query multiplication. Efficient Transformers mostly use approximate message passing which results in performance deterioration across tasks. Lightweight Convolution~\cite{wu2019pay}, Involution~\cite{li2021involution}, Synthesizer~\cite{tay2021synthesizer}, and MUSE~\cite{zhao2019muse} explored the relationship between Depthwise Convolution and Transformer. Our \model unification performs global and local information exchange simultaneously using a mixture affinity matrix, while \modellight switches off the dynamic affinity matrix for high resolution feature maps to reduce computation. Although switching off the dynamic affinity matrix slightly hinders classification performance, \modellight still provides effective and efficient generalization to downstream tasks compared with popular backbones such as ViT and ResNet.

\noindent \textbf{Transformers for Vision.}
Transformers enable high degrees of parallelism and are able to capture long-range dependencies in the input. Thus Transformers have gradually surpassed other architectures such as  CNN~\cite{lecun1998gradient} and RNN~\cite{hochreiter1997long} on image~\cite{dosovitskiy2020image,carion2020end,zhang2021rest}, audio~\cite{baevski2020wav2vec}, multi-modality~\cite{gao2019dynamic,geng2020character,geng2020dynamic}, and language understanding~\cite{devlin2018bert}. In computer vision, Non-local Neural Network~\cite{wang2018non} has been proposed to capture long range interactions to compensate for the local information captured by CNNs and used for object detection~\cite{hu2018relation} and semantic segmentation~\cite{fu2019dual,huang2019ccnet,zhu2019asymmetric,yuan2019object}. However, these methods use Transformers as a refinement module instead of treating the transformer as a first-class citizen. ViT~\cite{dosovitskiy2020image} introduces the first pure Transformer model into computer vision and surpasses CNNs with large scale pretraining on the non publicly available JFT dataset. DeiT~\cite{touvron2020training} trains ViT from scratch on ImageNet-1k and achieve better performance than CNN counterparts. DETR~\cite{carion2020end} uses Transformer as an encoder and decoder architecture for designing the first end-to-end object detection system. Taming Transformer~\cite{esser2020taming} use Vector Quantization~\cite{oord2017neural} GAN and GPT~\cite{radford2015unsupervised} for high quality high-resolution image generation. Motivated by the success of DETR on object detection, Transformers have been applied widely on tasks such as semantic segmentation~\cite{zheng2020rethinking}, pose estimation~\cite{yang2020transpose}, trajectory estimation~\cite{liu2021multimodal}, 3D representation learning and self-supervised learning with MOCO v3~\cite{chen2021empirical} and DINO~\cite{caron2021emerging}. ProTo~\cite{zhao2021proto} verify the effective of transformer on reasoning tasks.

\section{Methods}

In this section we first provide a generalized view of neighborhood/context aggregation modules commonly employed in present neural networks. Then we revisit three major architectures -- Transformer~\cite{vaswani2017attention}, Depthwise Convolution~\cite{chollet2017xception} and the recently proposed MLP-Mixer~\cite{tolstikhin2021mlpmixer}, and show that they are special cases of our generalized view. We then present our \model module in Sec~\ref{approach:container} and its efficient version -- \modellight in Sec~\ref{approach:container_light}.

\subsection{Contextual Aggregation for Vision}
\label{approach:context}

Consider an input image $X \in \mathbb{R}^{C \times H \times W}$, where $C$ and $H \times W$ denote the channel and spatial dimensions of the input image, respectively. The input image is first flattened to a sequence of tokens $\{X_i \in \mathbb{R}^C \vert i = 1, \ldots, N\}$, where $N = HW$ and input to the network. Vision networks typically stack multiple building blocks with residual connections \cite{he2016deep}, defined as
\begin{equation}
    \mathbf{Y} = \mathcal{F}(\mathbf{X},\{\mathbf{W}_i\}) + \mathbf{X}.
    \label{eq:res}
\end{equation}
Here, $\mathbf{X}$ and $\mathbf{Y}$ are the input and output vectors of the layers considered, and $\mathbf{W}_i$ represents the learnable parameters. $\mathcal{F}$ determines how information across $\mathbf{X}$ is aggregated to compute the feature at a specific location. We first define an affinity matrix $\mathcal{A} \in \mathbb{R}^{N \times N}$ that represents the neighborhood for contextual aggregation. Equation~\ref{eq:res} can be re-written as:  
\begin{equation}
    \mathbf{Y} = (\mathcal{A} \mathbf{V}) \mathbf{W}_1 + \mathbf{X}, \label{eq:single_affinity}
\end{equation}
where $\mathbf{V} \in \mathbb{R}^{N \times C}$ is a transformation of $\mathbf{X}$ obtained by a linear projection $\mathbf{V} = \mathbf{X} \mathbf{W}_2$. $\mathbf{W}_1$ and $\mathbf{W}_2$ are the learnable parameters.
$\mathcal{A}_{ij}$ is the affinity value between $X_i$ and $X_j$. 
Multiplying the affinity matrix with $\mathbf{V}$ propagates information across features in accordance with the affinity values. 

The modeling capacity of such a context aggregation module can be increased by introducing multiple affinity matrices, allowing the network to have several pathways to contextual information across $\mathbf{X}$. 
Let $\{\mathbf{V}^i \in \mathbb{R}^{N \times \frac{C}{M}} \vert i=1,\ldots,M\}$ be slices of $\mathbf{V}$, where $M$ is the number of affinity matrices, also referred to as the number of heads. The multi-head version of Equation~\ref{eq:single_affinity} is
\begin{equation}
    \mathbf{Y} = \operatorname{Concat}(\mathcal{A}_1 \mathbf{V}_1, \ldots, \mathcal{A}_M \mathbf{V}_M) \mathbf{W}_2 + \mathbf{X},
    \label{eq:multi_affinity}
\end{equation}
where $\mathcal{A}_m$ denotes the affinity matrix in each head. Different $A_m$ can potentially capture different relationships within the feature space and thus increase the representation power of contextual aggregation compared with a single-head version. 
Note that only spatial information is propagated during contextual aggregation using the affinity matrices; cross-channel information exchange does not occur within the affinity matrix multiplication, and that there is no non-linear activation function.



\subsection{The Transformer, Depthwise Convolution and MLP-Mixer}
\label{approach:threeblocks}

Transformer~\cite{vaswani2017attention}, depthwise convolution~\cite{kaiser2017depthwise} and the recently proposed MLP-Mixer~\cite{tolstikhin2021mlpmixer} are three distinct building blocks used in computer vision. Here, we show that they can be represented within the above context aggregation framework, by defining different types of affinity matrices.

\paragraph{Transformer.} In the self-attention mechanism in Transformers, the affinity matrix is modelled by the similarity between the projected query-key pairs. With $M$ heads, the affinity matrix in head $m$, $\mathcal{A}_m^{sa}$ can be written as
\begin{equation}
    \mathcal{A}_m^{sa} = \operatorname{Softmax}({\mathbf{Q}_m \mathbf{K}_m^{T}} / {\sqrt{{C} / {M}}}),
\end{equation}
where $\mathbf{K}_m, \mathbf{Q}_m$ are the corresponding key, query in head $m$, respectively.
The affinity matrix in self-attention is dynamically generated and can capture instance level information. However, this introduces quadratic computational, which requires heavy computation for high resolution feature. 



\paragraph{Depthwise Convolution.} 
The convolution operator fuses both spatial and channel information in parallel. This is different from the contextual aggregation block defined above. However, depthwise convolution \cite{kaiser2017depthwise} which is an extreme case of group convolution performs disentangled convolution. Considering the number of the heads from the contextual aggregation block to be equal to the channel size $C$, we can define the convolutional affinity matrix given the 1-d kernel ${Ker \in \mathbb{R}^{C \times 1 \times k}}$:
\begin{align}
    \mathcal{A}_{mij}^{conv} = \left\{
        \begin{array}{ll}
            Ker[m,0,\rvert i-j \lvert]  \quad &\lvert i - j  \rvert \leq k\\\
            0  \quad &\lvert i - j  \rvert > k 
        \end{array}
    \right., \label{eq:static}
\end{align}
where $\mathcal{A}_{mij}$ is the affinity value between $X_i$ and $X_j$ on head $m$. In contrast with the affinity matrix obtained from self-attention whose value is conditioned on the input feature, the affinity values for convolution are static -- they do not depend on the input features, sparse -- only involves local connections and shared across the affinity matrix. 
%

\paragraph{MLP-Mixer} 
The recently proposed MLP-Mixer~\cite{tolstikhin2021mlpmixer} does not rely on any convolution or self-attention operator. The core of MLP-Mixer is the transposed MLP operation,
which can be denoted as $\mathbf{X} = \mathbf{X} + (\mathbf{V}^T \mathbf{W}_{MLP})^T$. We can define the affinity matrix as
\begin{align}
    \mathcal{A}^{mlp} = (\mathbf{W}_{MLP})^T, \label{eq:residual}
\end{align}
where $\mathbf{W}_{MLP}$ represents the learnable parameters. This simple equation shows that the transposed-MLP operator is a contextual aggregation operator on a single feature group with a dense affinity matrix. Comparing with self-attention and depthwise convolution, the transpose-MLP affinity matrix is static, dense and with no parameter sharing. 



The above simple unification reveals the similarities and differences between Transformer, depthwise convolution and MLP-Mixer. Each of these building blocks can be obtained by different formulating different affinity matrices. This finding leads us to create a powerful and efficient building block for vision tasks -- the \model. 

\subsection{The \model Block}
\label{approach:container}
As detailed in Sec~\ref{approach:threeblocks}, previous architectures have employed either static or dynamically generated affinity matrices -- each of which provides its unique set of advantages and features. Our proposed building block named \model, combines both types of affinity matrices via a learnable parameter. The single head \model is defined as:
\begin{align}
    \mathbf{Y} &= ((\alpha \overbrace{ \mathcal{A}(\mathbf{X})}^{Dynamic} + \beta \overbrace{\mathcal{A}}^{Static})V) W_2 + \mathbf{X}
\end{align}
$\mathcal{A}(\mathbf{X})$ is dynamically generated from $\mathbf{X}$ while $\mathcal{A}$ is a static affinity matrix. We now present a few special cases of the \model block. In the following, $\mathcal{L}$ denotes a learnable parameter.

\begin{itemize}[leftmargin=*]
    \item $\alpha = 1$, $\beta = 0$, $\mathcal{A}(x)=\mathcal{A}^{sa}$: A vanilla Transformer block with self-attention (denoted $sa$).
    \item $\alpha = 0$, $\beta = 1$, $M=C$, $\mathcal{A} = \mathcal{A}^{conv}$: A depthwise convolution block. In depthwise convolution, each channel has a different static affinity matrix. When $M \neq C$, the resultant block can be considered a Multi-head Depthwise Convolution block (MH-DW). MH-DW shares kernel weights. 
    \item $\alpha = 0$, $\beta = 1$, $M=1$, $\mathcal{A} = \mathcal{A}^{mlp}$: An MLP-Mixer block. When $M \neq 1$, we name the module Multi-head MLP (MH-MLP). MH-MLP splits channels into $M$ groups and performs independent transposed MLP to capture diverse static token relationships.    
    \item $\alpha = \mathcal{L}$, $\beta = \mathcal{L}$, $\mathcal{A}(x)=\mathcal{A}^{sa}$, $\mathcal{A} = \mathcal{A}^{mlp}$: This \model block fuses dynamic and static information, but the static affinity resembles the MLP-Mixer matrix. We call this block \modelpam (Pay Attention to MLP).
    \item $\alpha = \mathcal{L}$, $\beta = \mathcal{L}$, $\mathcal{A}(x)=\mathcal{A}^{sa}$, $\mathcal{A} = \mathcal{A}^{conv}$: This \model block fuses dynamic and static information, but the static affinity resembles the depthwise convolution matrix. This static affinity matrix contains a locality constraint which is shift invariant, making it more suitable for vision tasks. This is the default configuration used in our experiments. 
\end{itemize}

The \model\ block is easy to implement and can be readily swapped into an existing neural network. The above versions of \model provide variations on the resulting architecture and its performance and exhibit different advantages and limitations. The computation cost of a \model block is the same as a vanilla Transformer since the static and dynamic matrices are linearly combined.

\subsection{The \model network architecture}
\label{approach:container_net}

We now present a base architecture used in our experiments. The unification of past works explained above allows us to easily compare self-attention, depthwise convolution, MLP and multiple variations of the \model block, and we perform these comparison using a consistent base architecture.

Motivated by networks in past works~\cite{he2016deep, wang2021pyramid}, our base architecture contains 4 stages. In contrast to ViT/DeiT which down-sample the image to a low resolution and keep this resolution constant, each stage in our architecture down-samples the image resolution gradually. Gradually down-sampling can retain image details, which is important for downstream tasks such as segmentation and detection. Each of the 4 stages contains a cascade of blocks. Each block contains two sub-modules, the first to aggregate spatial information (named spatial aggregation module) and the second to fuse channel information (named feed-forward module). In this paper, the channel fusion module is fixed to a 2-layer MLP as proposed in \cite{vaswani2017attention}. Designing a better spatial aggregation module is the main focus of this paper. 
The 4 stages contain 2, 3, 8 and 3 blocks respectively. Each stage uses patch embeddings which fuse spatial patches of size $p \times p$ into a single vector. For the 4 stages, the values of $p$ are 4,4,2,2 respectively.
The feature dimension within a stage remains constant -- and is set to 128, 256, 320, and 512 for the four stages. This base architecture augmented with the \model block results in a similar parameter size as DeiT-S~\cite{touvron2020training}.


\subsection{The \modellight network}
\label{approach:container_light}
We also present an efficient version known as \modellight which uses the same basic architecture as \model, but switches off the dynamic affinity matrix in the first 3 stages. The absence of the computation heavy dynamic attention at the early stages of computation help efficiently scale the model to process large image resolutions and achieve superior performance on downstream tasks such as detection and instance segmentation. 



\vspace{-4mm}
\begin{align}
    \mathcal{A}_m^{\modellight} = \left\{
        \begin{array}{ll}
            \mathcal{A}_m^{conv}  \quad  &Stage = 1, 2, 3\\\
            \alpha \mathcal{A}_m^{sa} + \beta \mathcal{A}_m^{conv}  \quad  &Stage = 4 
        \end{array}
    \right., \label{eq:static}
\end{align}
$\alpha$ and $\beta$ are learnable parameters. In network stage 1, 2, 3, \modellight will switch off $\mathcal{A}_m^{sa}$.

\section{Experiments}
\label{sec:exp}

\newcolumntype{d}{>{\columncolor{LightCyan}}c}
\newcolumntype{s}{>{\columncolor{LightCyan}}r}
\begin{table}
\setlength\tabcolsep{2pt}
    \centering
    \begin{tabular}{|c|c|crrr|cc|}
        \hline
        
        Family & Network & Top-1 Acc & Params & FLOPs & Throughput & Input dim & NAS\\[0.5ex]
        \hline \hline
        
        \multirow{11}{*}{CNN} & ResNet-50~\cite{he2016deep} & 78.5 & 25.6\mym & 4.1\myg & 1250.3 & $224^2$ & \xmark\\
        
        & ResNet-101~\cite{he2016deep} & 79.8 & 44.7\mym  & 7.9\myg &753.7& $224^2$ & \xmark \\
        \cline{2-8}
        
        & Xception71~\cite{chollet2017xception} & 79.9 & 42.3\mym  & \na & 423.5 & $299^2$ & \xmark\\
        \cline{2-8}
        \rowcolor{LightCyan}
        & RegNetY-4G~\cite{radosavovic2020designing} & 80.0 &21\mym&4.0\myg&1156.7& $224^2$ & \cmark\\
        
        & RegNetY-8G~\cite{radosavovic2020designing} & 81.7 &39\mym&8.0\myg&591.6&$224^2$ & \cmark\\
        
        & RegNetY-16G~\cite{radosavovic2020designing} & 82.9 &84\mym&16.0\myg&334.7&$224^2$ & \cmark\\
        \cline{2-8}
        \rowcolor{LightCyan}
        & EfficientNet-B3~\cite{tan2019efficientnet}&81.6&12\mym&1.8\myg&732.1 &$300^2$ & \cmark\\
        \rowcolor{LightCyan}
        & EfficientNet-B4~\cite{tan2019efficientnet}&82.9&19\mym&4.2\myg&349.4 &$380^2$ & \cmark\\
        
        & EfficientNet-B5~\cite{tan2019efficientnet}&83.6&30\mym&9.9\myg&169.1 &$456^2$ & \cmark\\
        
        & EfficientNet-B6~\cite{tan2019efficientnet}&84.0&43\mym&19.0\myg&96.9 &$528^2$ & \cmark\\
        
        & EfficientNet-B7~\cite{tan2019efficientnet}&84.3&66\mym&37.0\myg&55.1 &$600^2$ & \cmark\\
        \hline
        
        \multirow{14}{*}{Transformer} & ViT-B/16~\cite{dosovitskiy2020image} & 77.9 & 86\mym  & 55.4\myg & 85.9 &$384^2$ & \xmark\\
        
        & ViT-L/16~\cite{dosovitskiy2020image}  & 76.5 & 307\mym & 190.7\myg & 27.3 & $384^2$ & \xmark  \\
        \cline{2-8}
        \rowcolor{LightCyan}
        & DeiT-S~\cite{touvron2020training} & 79.9& 22.1\mym  &4.6\myg &940.4& $224^2$ & \xmark\\
        
        & DeiT-B~\cite{touvron2020training} & 81.8 & 86\mym  &17.5\myg & 292.3& $224^2$ & \xmark\\
%
        
        \cline{2-8}
        \rowcolor{LightCyan}
        & PVT-T~\cite{wang2021pyramid}&75.1&13.2\mym&1.9\myg&\na  &$224^2$ & \xmark\\
        \rowcolor{LightCyan}
        & PVT-S~\cite{wang2021pyramid}&79.8&24.5\mym&3.8\myg&\na &$224^2$ & \xmark\\
        
        & PVT-Medium~\cite{wang2021pyramid}&81.2&44.2\mym&6.7\myg&\na &$224^2$ & \xmark\\
        
        & PVT-L~\cite{wang2021pyramid}&81.7&61.4\mym&9.8\myg&\na &$224^2$ & \xmark\\
        \cline{2-8}
        \rowcolor{LightCyan}
        & ViL-T~\cite{Zhang2021MultiScaleVL}&76.3&6.7\mym&1.3\myg&\na &$224^2$ & \xmark\\
        \rowcolor{LightCyan}
        & ViL-S~\cite{Zhang2021MultiScaleVL}&82.0&24.6\mym&4.9\myg&\na &$224^2$ & \xmark\\
        \cline{2-8}
        \rowcolor{LightCyan}       
        & Swin-T~\cite{liu2021swin}& 81.3&29\mym&4.5\myg&755.2&$224^2$ & \xmark\\
        
        & Swin-S~\cite{liu2021swin}& 83.0&50\mym&8.7\myg&436.9&$224^2$ & \xmark\\
        
        & Swin-B~\cite{liu2021swin}& 83.3&88\mym&15.4\myg&278.1&$224^2$ & \xmark\\
        \hline
        
        \multirow{2}{*}{MLP} & Mixer-B/16~\cite{tolstikhin2021mlpmixer} & 76.4 & 79\mym &\na &\na &$224^2$ & \xmark\\
        \cline{2-8}
        
        & ResMLP-24~\cite{touvron2021resmlp}&79.4& 30\mym& 6.0\myg&715.4  &$224^2$ & \xmark\\        
        \hline

        \multirow{1}{*}{Hybrid} & ConvViT~\cite{d2021convit} & 81.3  & 27\mym &5.4\myg &\na &$224^2$ & \xmark\\
        \cline{2-8}
        \rowcolor{LightCyan} 
        & BoT-S1-50~\cite{srinivas2021bottleneck} & 79.1  & 20.8\mym &4.3\myg &\na &$224^2$ & \xmark \\
        & BoT-S1-59~\cite{srinivas2021bottleneck} & 81.7  & 33.5\mym &7.3\myg &\na &$224^2$ & \xmark \\
        \hline
        \rowcolor{LightCyan}
        Container & \model & 82.7 & 22.1\mym &8.1\myg&347.8& $224^2$ & \xmark\\
        \cline{2-8}
        \rowcolor{LightCyan}
        (Ours) & \modellight & 82.0 &20.0\mym & 3.2\myg&1156.9& $224^2$ & \xmark\\
        \hline
    \end{tabular}
    \vspace{6mm}
    \caption{ImageNet~\cite{deng2009imagenet} Top-1 accuracy comparison for CNN, Transformer, MLP, Hybrid and Container models. Throughput (images/s) is not reported in all papers (noted as \na). \colorbox{LightCyan}{Models that have fewer parameters than \model or upto 10\% more parameters are highlighted.}}
    \label{tab:imagenet_top1}
\end{table}

We now present experiments with \model for ImageNet and with \modellight for the tasks of object detection, instance segmentation and self-supervised learning. We also present appropriate baselines. Please see the appendix for details of the models, training and setup.

\subsection{ImageNet Classification}
\label{sec:exp_imagenet}

\textbf{Top-1 Accuracy.}
Table~\ref{tab:imagenet_top1} compares several highly performant models within the CNN, Transformer, MLP, Hybrid and our proposed \model families. \model and \modellight outperform the pure Transformer models ViT~\cite{dosovitskiy2020image} and DeiT~\cite{touvron2020training} despite far fewer parameters. They outperform PVT~\cite{wang2021pyramid} which employ a hierarchical representation similar to our base architecture. They also outperform the recently published state-of-the-art SWIN~\cite{liu2021swin} (they outperform Swin-T which has more parameters). The best performing models continue to be from the EfficientNet~\cite{tan2019efficientnet} family, but we note that EfficientNet~\cite{tan2019efficientnet} and RegNet~\cite{radosavovic2020designing} apply an extensive neural architecture search, which we do not. Finally note that \modellight not only achieves a high accuracy but does so at lower FLOPs and much faster throughput than models with comparable capacities.

The \model framework allows us to easily reproduce past architectures but also to create effective extensions over past work (outlined in Sec~\ref{approach:container}), several of which are compared in Table~\ref{tab:imagenet_ablation}. H-DeiT-S is a hierarchical version of DeiT-S obtained by simply using $\calA^{sa}$ within our hierarchical architecture and provides 1.2 gain. Conv-3 (naive convolution (conv) with $3 \times 3$ kernel) aggregates spatial and channel information, where as Group Conv-3 splits input features and performs convs using different kernels -- it is cheaper and more effective. When group size $=$ channel dim., we get depth-wise conv. DW-3 is a depthwise convs with 3 by 3 kernel that only aggregates spatial information. Channel information is fused using $1 \times 1$ convs. MH-DW-3 is a multi-head version of DW-3. MH-DW-3 shares kernel parameters within the same group. With fewer kernels, MH-DW-3 achieves comparable performance with DW-3. MLP is an implementation of transposed MLP for spatial propagation. MLP-LR stands for MLP with low-rank decompostion. MLP-LR provides better performance with fewer parameters. MH-MLP-LR adds a multi-head mechanism over MLP-LR and provides further improvements. In contrast to the original MLP-Mixer~\cite{tolstikhin2021mlpmixer}, we do not add any non-linearity like GELU into \model as is specified in the contextual aggregation equation. 


\begin{table}
\setlength\tabcolsep{2pt}
    \centering
    \begin{tabular}{|c|cc|ccccc|}
        \hline
        Method & Top-1 Acc & Params & $\alpha$ & $\beta$ & $\frac{C}{M}$ & $\calA^{dynamic}$ & $\calA^{static}$\\[0.5ex]
        \hline \hline
        H-DeiT-S & 81.0& 22.1\mym  & 1 & 0 & 32 & $\calA^{sa}$ & N/A\\
        \hline
        Conv-3 & 79.6 & 33.8\mym &\na&\na&\na&\na & \na\\
        Group Conv-3 & 79.7 & 20.5\mym &\na&\na&\na & \na &\na\\
        DW-3 & 80.1 &18.7\mym  & 0 & 1 & 1 & \na & $\calA^{conv}$ \\
        MH-DW-3 & 79.9 &18.6\mym & 0 & 1 & 32 & \na & $\calA^{conv}$  \\
        \hline
        MLP &77.5 & 50.9\mym& 0 & 1 & C & \na & $\calA^{mlp}$ \\
        MLP-LR &78.9 &36.5\mym & 0 & 1 & C & \na & $\calA^{mlp}$ \\
        MH-MLP-LR &79.6 &41.6 \mym & 0 & 1 & 32 & \na & $\calA^{mlp}$ \\
        \hline
        \model  &82.7 &22.1\mym& $\calL$ & $\calL$ & 32 & $\calA^{sa}$ &$\calA^{conv}$\\
        \modellight  &82.0 &20.0\mym & $\calL$ & $\calL$ & 32 & $\calA^{sa}$ &$\calA^{conv}$\\
        \hline
    \end{tabular}
    \vspace{6mm}
    \caption{ImageNet accuracies for architecture variations (with convolutions, self-attention and MLP) enabled within the \model framework. As per our notation, $C$: num channels, $M$: num heads, $C/M$: head dimension. See Sec~\ref{approach:container} and ~\ref{sec:exp_imagenet} for notation and model details.}
    \label{tab:imagenet_ablation}
\end{table}

\textbf{Data Efficiency.}
\begin{wraptable}[8]{R}{0.4\linewidth}
\vspace{-4mm}
\footnotesize
\setlength\tabcolsep{2pt}
    \centering
    \begin{tabular}{|c|c|c|}
        \hline
        Data ratio  & \modellight & DeiT \\
        \hline
        100 \% & 82.0 (+2.1) & 79.9\\
        80 \% & 81.1 (+2.6)& 78.5 \\
        50 \%  & 78.8 (+4.8) & 74.0 \\
        10 \% & 61.8 (+22.5)& 39.3 \\
        \hline
    \end{tabular}
    \vspace{2pt}
    \caption{ImageNet Top-1 Acc for \modellight and DeiT-S with varying training sizes.}
    \label{tab:data_efficiency}
\end{wraptable}
\modellight has a built-in shift-invariance and parameter sharing mechanism. As a result it is more data efficient in comparison to DeiT~\cite{touvron2020training}. Table~\ref{tab:data_efficiency} shows that at the low data regime of 10\%, \modellight outperforms DeiT by a massive 22.5 points.

\begin{figure}[]
        \begin{center}
        \includegraphics[width=\linewidth]{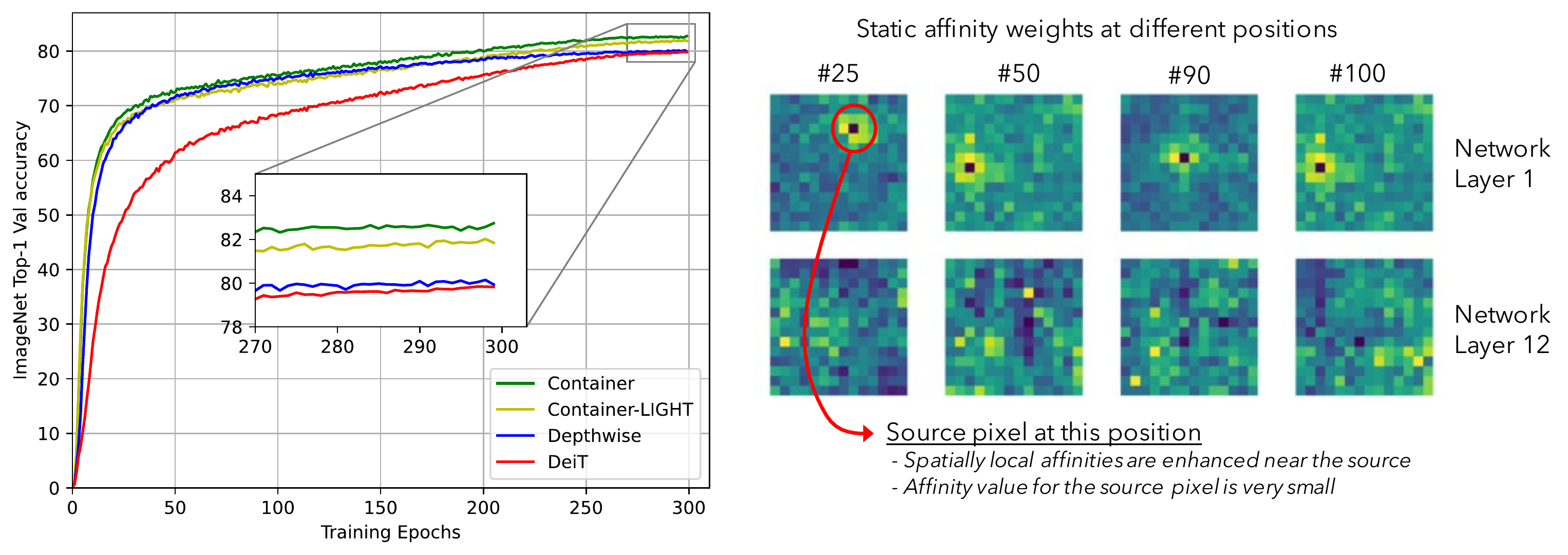}
        \end{center}
        \caption{\textbf{(left)} Convergence speed comparison between \model, \modellight, Depthwise conv and DeiT. \textbf{(right)} Visualization of the static affinity weights at different positions and layers. Layer 1 displays the emergence of local affinities (resembling convolutions).}
        \label{fig:convergence_viz}
\end{figure}

\textbf{Convergence Speed.}
Figure~\ref{fig:convergence_viz} (left) compares the convergence speeds of the two \model variants with a CNN and Transformer (DeiT)~\cite{touvron2020training}. The inductive biases in the CNN enable it to converge faster than DeiT~\cite{touvron2020training}, but they eventually perform similarly at 300 epochs, suggesting that dynamic, long range context aggregation is powerful but slow to converge. \model combines the best of both and provides accuracy improvements with fast convergence. \modellight converges as fast with a slight accuracy drop.

\textbf{Emergence of locality.}
Within our \model framework, we can easily add a static affinity matrix to the DeiT architecture. This simple change (1 line of code addition), can provide a +0.5 Top-1 improvement from 79.9\% to 80.4\%. This suggests that static and dynamic affinity matrices provide complementary information. As noted in Sec~\ref{approach:container}, we name this \modelpam. 

It is interesting to visualize the learnt static affinities at different network layers. Figure~\ref{fig:convergence_viz} (right) displays these for 2 layers. Each matrix represents the static affinities for a single position, reshaped to a 2-d grid to resemble the landscape of the neighboring regions. 
Within Layer 1,  we interestingly observe the emergence of local operations via the enhancement of affinity values next to the source pixel (location). These are akin to convolution operations. Furthermore, the affinity value for the source pixel is very small, i.e. at each location, the context aggregator does not use its current feature. We hypothesize that this is a result of the residual connection~\cite{he2016deep}, thereby alleviating the need to include the source feature within the context. Note that in contrast to dynamic affinity, the learnt static matrix is shared for all input images. Notice that Layer 12 displays a more global affinity matrix without any specific interpretable local patterns.

\subsection{Detection with RetinaNet}
\begin{table*} \footnotesize
\setlength\tabcolsep{2pt}
    \centering
    \begin{tabular}{c|cc|c|cc|c|cc|cc|cccc}
        \toprule
        & \multicolumn{8}{l|}{\makecell[c]{Mask R-CNN}} & \multicolumn{6}{l}{\makecell[c]{RetinaNet}} \\ \midrule
        Method &  \#P&FLOPs&  $AP^b$ & $AP^b_{50}$ & $AP^b_{75}$ & $AP^m$ & $AP^m_{50}$ & $AP^m_{75}$ & \#P&FLOPs& mAP & AP$_S$ & AP$_M$ & AP$_L$\\
        \midrule
        ResNet50~\cite{he2016deep} & 44.2& 180\myg& 38.2 &58.8& 41.4&34.7 &55.7 & 37.2 &37.7 &239\myg & 36.5&20.4& 40.3&48.1\\
        ResNet101~\cite{he2016deep}& 63.2& 259\myg&40.0 &60.5& 44.0&36.1 & 57.5 & 38.6 &56.7&319\myg  & 38.5&21.7& 42.8&50.4\\
        X-101-32~\cite{xie2017aggregated}&62.8 &259\myg & 41.9 & 62.5 & 45.9 & 37.5 & 59.4 & 40.2 &56.4&319\myg & 39.9 & 22.3 & 44.2 & 52.5 \\
        X-101-64~\cite{xie2017aggregated}& 101.9 & 424\myg& 42.8 & 63.8 & 47.3 & 38.4 & 60.6 & 41.3 & 95.5&483\myg & 41.0& 23.9 & 45.2 & 54.0\\
        \midrule
        PVT-S~\cite{wang2021pyramid} & 44.1 &245\myg & 40.4 & 62.9 & 43.8 & 37.8 & 60.1 & 40.3 &34.2& 226\myg& 40.4& 25.0 & 42.9 & 55.7 \\  
        ViL-S~\cite{wang2021pyramid} & 45.0&174\myg& 41.8 & 64.1 & 45.1 & 38.5 & 61.1 & 41.4 & 35.6&252\myg& 41.6& 24.9 & 44.6 & 56.2  \\ 
        SWIN-T~\cite{liu2021swin} & 48.0&267\myg& 43.7 & 66.6 & 47.4 & 39.8 & 63.6 & 42.7 & 385 &244\myg & 41.5 & 26.4 & 45.1 & 55.7\\ 
        ViL-M~\cite{wang2021pyramid} & 60.1&261\myg& 43.4 & 65.9 & 47.0 & 39.7 & 62.8 & 42.1 & 50.7&338\myg& 42.9& 27.0 & 46.1 & 57.2\\
        ViL-B~\cite{wang2021pyramid} & 76.1&365\myg& 45.1 & 67.2 & 49.3 & 41.0 & 64.3 & 44.2 & 66.7&443\myg& 44.3 & 28.9 & 47.9 & 58.3\\ 
        \midrule
        BoT50~\cite{srinivas2021bottleneck}  & 39.5&\na& 39.4 & 60.3 & 43.0 & 35.3 & 57 & 37.5 & \na & \na & \na & \na & \na & \na\\
        BoT50-(6x)~\cite{srinivas2021bottleneck} & 39.5&\na& 43.7 & 64.7 & 47.9 & 38.7 & 61.8 & 41.1 & \na & \na & \na & \na & \na & \na\\
        \midrule
        \scriptsize{\modellight} &39.6 &237\myg &45.1&67.3 &49.5 &41.3 &64.2 &44.5  & 29.7&218\myg & 43.8 &27.4 &47.5 &58.5\\
        \bottomrule
    \end{tabular}
    \vspace{2pt}
    \caption{Comparing the \modellight backbone with several previous methods at the tasks of object detection and instance segmentation using the Mask-RCNN and RetinaNet networks.}
    \label{tab:detection_segmentation}
\end{table*}

Since the attention complexity for \modellight is linear at high image resolutions (initial layers) and then quadratic, it can be employed for downstream tasks such as object detection which usually require high resolution feature maps. Table~\ref{tab:detection_segmentation} compares several backbones applied to the RetinaNet detector~\cite{lin2017focal} on the COCO dataset~\cite{lin2014microsoft}. Compared to the popular ResNet-50~\cite{he2016deep}, \modellight achieves 43.8 mAP, an improvement of 7.0, 7.2 and 10.4 on $AP_S$, $AP_M$, and $AP_L$ with comparable parameters and cost. The significant increase for large objects shows the benefits of global attention via the dynamic global affinity matrix in our model. \modellight also surpasses the large convolution-based backbone X-101-64~\cite{xie2017aggregated} and pure Transformer models with similar number of parameters such as PVT-S~\cite{wang2021pyramid}, ViL-S~\cite{Zhang2021MultiScaleVL}, and SWIN-T~\cite{liu2021swin} by large margins. Compared to large Transformer backbones such as ViL-M~\cite{Zhang2021MultiScaleVL} and ViL-B~\cite{Zhang2021MultiScaleVL}, we achieve comparable performance with significantly fewer parameters and FLOPs. 

\subsection{Detection and Segmentation with Mask-RCNN}
Table~\ref{tab:detection_segmentation} also compares several backbones for detection and instance segmentation using the Mask R-CNN network~\cite{he2017mask}. As with the findings for RetinaNet~\cite{lin2017focal}, \modellight outperforms convolution and Transformer based approaches such as ResNet~\cite{he2016deep}, X-101~\cite{xie2017aggregated}, PVT~\cite{wang2021pyramid}, ViL~\cite{Zhang2021MultiScaleVL} and recent state-of-the-art SWIN-T~\cite{liu2021swin} and the recent hybrid approach BoT~\cite{srinivas2021bottleneck}. It obtains comparable numbers to the much larger ViL-B~\cite{Zhang2021MultiScaleVL}.

\subsection{Detection with DETR}
\begin{wraptable}[8]{R}{0.5\linewidth}
\vspace{-10mm}
\footnotesize
\setlength\tabcolsep{2pt}
    \centering
    \begin{tabular}{|c|c|}
        \hline
        Method   & mAP \\
        \hline
        DETR-ResNet50~\cite{carion2020end} & 32.3\\
        DETR-\modellight & 38.9 \\
        \hline
        DDETR w/o multi-scale-ResNet50~\cite{zhu2020deformable}  &  39.3 \\
        DDETR w/o multi-scale-\modellight  & 43.0 \\
        \hline
        SMCA w/o multi-scale-ResNet50~\cite{gao2021fast}  &  41.0 \\
        SMCA w/o multi-scale-\modellight  &  44.2 \\
        \hline
    \end{tabular}
    \vspace{2pt}
    \caption{\modellight and ResNet-50 backbones with DETR and variants for object detection.}
    \label{tab:detr}
\end{wraptable}
Table~\ref{tab:detr} shows that our model can consistently improve object detection performance compared to a ResNet-50~\cite{he2016deep} backbone (comparable parameters and computation) on end-to-end object detection using DETR~\cite{carion2020end}. We demonstrate large improvements with DETR~\cite{carion2020end}, DDETR~\cite{zhu2020deformable} as well as SMCA-DETR~\cite{gao2021fast}. See appendix for $AP^S$, $AP^M$, and $AP^L$ numbers. All models in table ~\ref{tab:detr} are trained using a 50 epochs schedule. 

\subsection{Self supervised learning}
\begin{wraptable}[6]{R}{0.45\linewidth}
\vspace{-8mm}
\footnotesize
\setlength\tabcolsep{2pt}
    \centering
    \begin{tabular}{c|ccccc}
        \toprule
        Epochs $\rightarrow$ &  20 &40&  60 & 80 & 100 \\
        \midrule
        DeiT ~\cite{caron2021emerging}  & 52.0& 63.3& 66.5 &68.9& 69.6\\
        \modellight &58.0 & 67.0 & 70.0 & 71.1 & 71.5 \\
        \bottomrule
    \end{tabular}
    \vspace{2pt}
    \caption{\modellight and DeiT on DINO self-supervised learning.}
    \label{tab:dino}
\end{wraptable}

We train DeiT~\cite{touvron2020training} and \modellight for 100 epochs at the self supervised task of visual representation learning using the DINO framework~\cite{caron2021emerging}. Table~\ref{tab:dino} compares top-10 kNN accuracy for both backbones at different epochs of training. \modellight significantly outperforms DeiT with large improvements initially demonstrating more efficient learning.

\section{Conclusion}
\label{sec:conclusion}

In this paper, we have shown that disparate architectures such as Transformers, depth-wise CNNs and MLP-based methods are closely related via an affinity matrix used for context aggregation. Using this view, we have proposed \model, a generalized context aggregation building block that combines static and dynamic affinity matrices using learnable parameters. Our proposed networks, \model and \modellight show superior performance at image classification, object detection, instance segmentation and self-supervised representation learning. We hope that this unified view can motivate future research in the design of effective and efficient visual backbones.

\textbf{Limitations}: \model is very effective at image classification but cannot be directly applied to high resolution inputs. The efficient version \modellight, can be used for a variety of tasks. However, its limitation is that it is partially hand-crafted -- the dynamic affinity matrix is switched off in the first 3 stages. Future work will address how to learn this using the task at hand.

\textbf{Negative societal impact}: This research does not have a direct negative societal impact. However, we should be aware that powerful neural networks, particularly image classification networks can be used for harmful applications like face and gender recognition.

\textbf{Disclosure of Funding}
This work was partially supported by the Shanghai Committee of Science and Technology, China (Grant No. 21DZ1100100 and 20DZ1100800).



{\small
\bibliographystyle{bib/ieee_fullname}
\bibliography{bib/egbib.bib}

\begin{thebibliography}{10}\itemsep=-1pt

\bibitem{Anderson2018BottomUpAT}
Peter Anderson, X. He, Chris Buehler, Damien Teney, Mark Johnson, Stephen
  Gould, and Lei Zhang.
\newblock Bottom-up and top-down attention for image captioning and visual
  question answering.
\newblock In {\em CVPR}, 2018.

\bibitem{baevski2020wav2vec}
Alexei Baevski, Henry Zhou, Abdelrahman Mohamed, and Michael Auli.
\newblock wav2vec 2.0: A framework for self-supervised learning of speech
  representations.
\newblock In {\em NeurIPS}, 2020.

\bibitem{beltagy2020longformer}
Iz Beltagy, Matthew~E Peters, and Arman Cohan.
\newblock Longformer: The long-document transformer.
\newblock {\em arXiv}, 2020.

\bibitem{brown2020language}
Tom Brown, Benjamin Mann, Nick Ryder, Melanie Subbiah, Jared~D Kaplan, Prafulla
  Dhariwal, Arvind Neelakantan, Pranav Shyam, Girish Sastry, Amanda Askell,
  Sandhini Agarwal, Ariel Herbert-Voss, Gretchen Krueger, Tom Henighan, Rewon
  Child, Aditya Ramesh, Daniel Ziegler, Jeffrey Wu, Clemens Winter, Chris
  Hesse, Mark Chen, Eric Sigler, Mateusz Litwin, Scott Gray, Benjamin Chess,
  Jack Clark, Christopher Berner, Sam McCandlish, Alec Radford, Ilya Sutskever,
  and Dario Amodei.
\newblock Language models are few-shot learners.
\newblock In {\em NeurIPS}, 2020.

\bibitem{carion2020end}
Nicolas Carion, Francisco Massa, Gabriel Synnaeve, Nicolas Usunier, Alexander
  Kirillov, and Sergey Zagoruyko.
\newblock End-to-end object detection with transformers.
\newblock In {\em ECCV}, 2020.

\bibitem{caron2021emerging}
Mathilde Caron, Hugo Touvron, Ishan Misra, Herv{\'e} J{\'e}gou, Julien Mairal,
  Piotr Bojanowski, and Armand Joulin.
\newblock Emerging properties in self-supervised vision transformers.
\newblock {\em arXiv}, 2021.

\bibitem{chen2021empirical}
Xinlei Chen, Saining Xie, and Kaiming He.
\newblock An empirical study of training self-supervised visual transformers.
\newblock {\em arXiv}, 2021.

\bibitem{chollet2017xception}
Fran{\c{c}}ois Chollet.
\newblock Xception: Deep learning with depthwise separable convolutions.
\newblock In {\em CVPR}, 2017.

\bibitem{choromanski2020rethinking}
Krzysztof Choromanski, Valerii Likhosherstov, David Dohan, Xingyou Song,
  Andreea Gane, Tamas Sarlos, Peter Hawkins, Jared Davis, Afroz Mohiuddin,
  Lukasz Kaiser, et~al.
\newblock Rethinking attention with performers.
\newblock In {\em ICLR}, 2021.

\bibitem{daras2020smyrf}
Giannis Daras, Nikita Kitaev, Augustus Odena, and Alexandros~G Dimakis.
\newblock Smyrf: Efficient attention using asymmetric clustering.
\newblock In {\em NeurIPS}, 2020.

\bibitem{d2021convit}
St{\'e}phane d'Ascoli, Hugo Touvron, Matthew Leavitt, Ari Morcos, Giulio
  Biroli, and Levent Sagun.
\newblock Convit: Improving vision transformers with soft convolutional
  inductive biases.
\newblock {\em arXiv}, 2021.

\bibitem{deng2009imagenet}
Jia Deng, Wei Dong, Richard Socher, Li-Jia Li, Kai Li, and Li Fei-Fei.
\newblock Imagenet: A large-scale hierarchical image database.
\newblock In {\em CVPR}, 2009.

\bibitem{devlin2018bert}
Jacob Devlin, Ming-Wei Chang, Kenton Lee, and Kristina Toutanova.
\newblock Bert: Pre-training of deep bidirectional transformers for language
  understanding.
\newblock In {\em NAACL}, 2019.

\bibitem{dosovitskiy2020image}
Alexey Dosovitskiy, Lucas Beyer, Alexander Kolesnikov, Dirk Weissenborn,
  Xiaohua Zhai, Thomas Unterthiner, Mostafa Dehghani, Matthias Minderer, Georg
  Heigold, Sylvain Gelly, et~al.
\newblock An image is worth 16x16 words: Transformers for image recognition at
  scale.
\newblock In {\em ICLR}, 2021.

\bibitem{esser2020taming}
Patrick Esser, Robin Rombach, and Bj{\"o}rn Ommer.
\newblock Taming transformers for high-resolution image synthesis.
\newblock In {\em CVPR}, 2021.

\bibitem{fu2019dual}
Jun Fu, Jing Liu, Haijie Tian, Yong Li, Yongjun Bao, Zhiwei Fang, and Hanqing
  Lu.
\newblock Dual attention network for scene segmentation.
\newblock In {\em CVPR}, 2019.

\bibitem{gao2019dynamic}
Peng Gao, Zhengkai Jiang, Haoxuan You, Pan Lu, Steven~CH Hoi, Xiaogang Wang,
  and Hongsheng Li.
\newblock Dynamic fusion with intra-and inter-modality attention flow for
  visual question answering.
\newblock In {\em Proceedings of the IEEE/CVF Conference on Computer Vision and
  Pattern Recognition}, pages 6639--6648, 2019.

\bibitem{gao2019multi}
Peng Gao, Haoxuan You, Zhanpeng Zhang, Xiaogang Wang, and Hongsheng Li.
\newblock Multi-modality latent interaction network for visual question
  answering.
\newblock In {\em ICCV}, 2019.

\bibitem{gao2021fast}
Peng Gao, Minghang Zheng, Xiaogang Wang, Jifeng Dai, and Hongsheng Li.
\newblock Fast convergence of detr with spatially modulated co-attention.
\newblock {\em arXiv}, 2021.

\bibitem{geng2020dynamic}
Shijie Geng, Peng Gao, Moitreya Chatterjee, Chiori Hori, Jonathan~Le Roux,
  Yongfeng Zhang, Hongsheng Li, and Anoop Cherian.
\newblock Dynamic graph representation learning for video dialog via
  multi-modal shuffled transformers.
\newblock {\em arXiv preprint arXiv:2007.03848}, 2020.

\bibitem{geng2020character}
Shijie Geng, Ji Zhang, Zuohui Fu, Peng Gao, Hang Zhang, and Gerard de Melo.
\newblock Character matters: Video story understanding with character-aware
  relations.
\newblock {\em arXiv preprint arXiv:2005.08646}, 2020.

\bibitem{he2017mask}
Kaiming He, Georgia Gkioxari, Piotr Doll{\'a}r, and Ross Girshick.
\newblock Mask r-cnn.
\newblock In {\em ICCV}, 2017.

\bibitem{he2016deep}
Kaiming He, Xiangyu Zhang, Shaoqing Ren, and Jian Sun.
\newblock Deep residual learning for image recognition.
\newblock In {\em CVPR}, 2016.

\bibitem{Heo2021RethinkingSD}
Byeongho Heo, Sangdoo Yun, Dongyoon Han, Sanghyuk Chun, Junsuk Choe, and
  Seong~Joon Oh.
\newblock Rethinking spatial dimensions of vision transformers.
\newblock {\em arXiv}, 2021.

\bibitem{hochreiter1997long}
Sepp Hochreiter and J{\"u}rgen Schmidhuber.
\newblock Long short-term memory.
\newblock {\em Neural computation}, 9(8), 1997.

\bibitem{howard2017mobilenets}
Andrew~G Howard, Menglong Zhu, Bo Chen, Dmitry Kalenichenko, Weijun Wang,
  Tobias Weyand, Marco Andreetto, and Hartwig Adam.
\newblock Mobilenets: Efficient convolutional neural networks for mobile vision
  applications.
\newblock {\em arXiv}, 2017.

\bibitem{hu2018relation}
Han Hu, Jiayuan Gu, Zheng Zhang, Jifeng Dai, and Yichen Wei.
\newblock Relation networks for object detection.
\newblock In {\em CVPR}, 2018.

\bibitem{hu2018squeeze}
Jie Hu, Li Shen, and Gang Sun.
\newblock Squeeze-and-excitation networks.
\newblock In {\em CVPR}, 2018.

\bibitem{huang2019ccnet}
Zilong Huang, Xinggang Wang, Lichao Huang, Chang Huang, Yunchao Wei, and Wenyu
  Liu.
\newblock Ccnet: Criss-cross attention for semantic segmentation.
\newblock In {\em ICCV}, 2019.

\bibitem{kaiser2017depthwise}
Lukasz Kaiser, Aidan~N Gomez, and Francois Chollet.
\newblock Depthwise separable convolutions for neural machine translation.
\newblock {\em arXiv}, 2017.

\bibitem{katharopoulos2020transformers}
Angelos Katharopoulos, Apoorv Vyas, Nikolaos Pappas, and Fran{\c{c}}ois
  Fleuret.
\newblock Transformers are rnns: Fast autoregressive transformers with linear
  attention.
\newblock In {\em ICML}, 2020.

\bibitem{kitaev2020reformer}
Nikita Kitaev, {\L}ukasz Kaiser, and Anselm Levskaya.
\newblock Reformer: The efficient transformer.
\newblock In {\em ICLR}, 2020.

\bibitem{krizhevsky2012imagenet}
Alex Krizhevsky, Ilya Sutskever, and Geoffrey~E Hinton.
\newblock Imagenet classification with deep convolutional neural networks.
\newblock {\em NeurIPS}, 2012.

\bibitem{lecun1998gradient}
Yann LeCun, L{\'e}on Bottou, Yoshua Bengio, and Patrick Haffner.
\newblock Gradient-based learning applied to document recognition.
\newblock {\em Proceedings of the IEEE}, 86(11), 1998.

\bibitem{li2021involution}
Duo Li, Jie Hu, Changhu Wang, Xiangtai Li, Qi She, Lei Zhu, Tong Zhang, and
  Qifeng Chen.
\newblock Involution: Inverting the inherence of convolution for visual
  recognition.
\newblock In {\em Proceedings of the IEEE/CVF Conference on Computer Vision and
  Pattern Recognition}, pages 12321--12330, 2021.

\bibitem{Li2021LocalViTBL}
Yawei Li, Kai Zhang, Jiezhang Cao, R. Timofte, and L. Gool.
\newblock Localvit: Bringing locality to vision transformers.
\newblock {\em arXiv}, 2021.

\bibitem{lin2017focal}
Tsung-Yi Lin, Priya Goyal, Ross Girshick, Kaiming He, and Piotr Doll{\'a}r.
\newblock Focal loss for dense object detection.
\newblock In {\em ICCV}, 2017.

\bibitem{lin2014microsoft}
Tsung-Yi Lin, Michael Maire, Serge Belongie, James Hays, Pietro Perona, Deva
  Ramanan, Piotr Doll{\'a}r, and C~Lawrence Zitnick.
\newblock Microsoft coco: Common objects in context.
\newblock In {\em ECCV}, 2014.

\bibitem{liu2021multimodal}
Yicheng Liu, Jinghuai Zhang, Liangji Fang, Qinhong Jiang, and Bolei Zhou.
\newblock Multimodal motion prediction with stacked transformers.
\newblock In {\em CVPR}, 2021.

\bibitem{liu2021swin}
Ze Liu, Yutong Lin, Yue Cao, Han Hu, Yixuan Wei, Zheng Zhang, Stephen Lin, and
  Baining Guo.
\newblock Swin transformer: Hierarchical vision transformer using shifted
  windows.
\newblock {\em arXiv}, 2021.

\bibitem{oord2017neural}
Aaron van~den Oord, Oriol Vinyals, and Koray Kavukcuoglu.
\newblock Neural discrete representation learning.
\newblock In {\em NeurIPS}, 2017.

\bibitem{radford2021learning}
Alec Radford, Jong~Wook Kim, Chris Hallacy, Aditya Ramesh, Gabriel Goh,
  Sandhini Agarwal, Girish Sastry, Amanda Askell, Pamela Mishkin, Jack Clark,
  Gretchen Krueger, and Ilya Sutskever.
\newblock Learning transferable visual models from natural language
  supervision.
\newblock {\em arXiv}, 2021.

\bibitem{radford2015unsupervised}
Alec Radford, Luke Metz, and Soumith Chintala.
\newblock Unsupervised representation learning with deep convolutional
  generative adversarial networks.
\newblock {\em arXiv}, 2015.

\bibitem{radosavovic2020designing}
Ilija Radosavovic, Raj~Prateek Kosaraju, Ross Girshick, Kaiming He, and Piotr
  Doll{\'a}r.
\newblock Designing network design spaces.
\newblock In {\em CVPR}, 2020.

\bibitem{rajpurkar2016squad}
Pranav Rajpurkar, Jian Zhang, Konstantin Lopyrev, and Percy Liang.
\newblock Squad: 100,000+ questions for machine comprehension of text.
\newblock {\em arXiv}, 2016.

\bibitem{simonyan2014very}
Karen Simonyan and Andrew Zisserman.
\newblock Very deep convolutional networks for large-scale image recognition.
\newblock In {\em ICLR}, 2015.

\bibitem{srinivas2021bottleneck}
Aravind Srinivas, Tsung-Yi Lin, Niki Parmar, Jonathon Shlens, Pieter Abbeel,
  and Ashish Vaswani.
\newblock Bottleneck transformers for visual recognition.
\newblock {\em arXiv}, 2021.

\bibitem{szegedy2015going}
Christian Szegedy, Wei Liu, Yangqing Jia, Pierre Sermanet, Scott Reed, Dragomir
  Anguelov, Dumitru Erhan, Vincent Vanhoucke, and Andrew Rabinovich.
\newblock Going deeper with convolutions.
\newblock In {\em CVPR}, 2015.

\bibitem{tan2019efficientnet}
Mingxing Tan and Quoc Le.
\newblock Efficientnet: Rethinking model scaling for convolutional neural
  networks.
\newblock In {\em ICML}, 2019.

\bibitem{Tan2021EfficientNetV2SM}
Mingxing Tan and Quoc~V. Le.
\newblock Efficientnetv2: Smaller models and faster training.
\newblock {\em arXiv}, 2021.

\bibitem{tay2021synthesizer}
Yi Tay, Dara Bahri, Donald Metzler, Da-Cheng Juan, Zhe Zhao, and Che Zheng.
\newblock Synthesizer: Rethinking self-attention for transformer models.
\newblock In {\em International Conference on Machine Learning}, pages
  10183--10192. PMLR, 2021.

\bibitem{tolstikhin2021mlpmixer}
Ilya Tolstikhin, Neil Houlsby, Alexander Kolesnikov, Lucas Beyer, Xiaohua Zhai,
  Thomas Unterthiner, Jessica Yung, Daniel Keysers, Jakob Uszkoreit, Mario
  Lucic, and Alexey Dosovitskiy.
\newblock Mlp-mixer: An all-mlp architecture for vision.
\newblock {\em arXiv}, 2021.

\bibitem{touvron2021resmlp}
Hugo Touvron, Piotr Bojanowski, Mathilde Caron, Matthieu Cord, Alaaeldin
  El-Nouby, Edouard Grave, Armand Joulin, Gabriel Synnaeve, Jakob Verbeek, and
  Herv{\'e} J{\'e}gou.
\newblock Resmlp: Feedforward networks for image classification with
  data-efficient training.
\newblock {\em arXiv}, 2021.

\bibitem{touvron2020training}
Hugo Touvron, Matthieu Cord, Matthijs Douze, Francisco Massa, Alexandre
  Sablayrolles, and Herv{\'e} J{\'e}gou.
\newblock Training data-efficient image transformers \& distillation through
  attention.
\newblock {\em arXiv}, 2020.

\bibitem{Vaswani2021ScalingLS}
Ashish Vaswani, Prajit Ramachandran, A. Srinivas, Niki Parmar, Blake~A.
  Hechtman, and Jonathon Shlens.
\newblock Scaling local self-attention for parameter efficient visual
  backbones.
\newblock {\em arXiv}, 2021.

\bibitem{vaswani2017attention}
Ashish Vaswani, Noam Shazeer, Niki Parmar, Jakob Uszkoreit, Llion Jones,
  Aidan~N Gomez, Lukasz Kaiser, and Illia Polosukhin.
\newblock Attention is all you need.
\newblock In {\em NeurIPS}, 2017.

\bibitem{vyas2020fast}
Apoorv Vyas, Angelos Katharopoulos, and Fran{\c{c}}ois Fleuret.
\newblock Fast transformers with clustered attention.
\newblock In {\em NeurIPS}, 2020.

\bibitem{wang2018glue}
Alex Wang, Amanpreet Singh, Julian Michael, Felix Hill, Omer Levy, and Samuel~R
  Bowman.
\newblock Glue: A multi-task benchmark and analysis platform for natural
  language understanding.
\newblock {\em arXiv}, 2018.

\bibitem{wang2020linformer}
Sinong Wang, Belinda Li, Madian Khabsa, Han Fang, and Hao Ma.
\newblock Linformer: Self-attention with linear complexity.
\newblock {\em arXiv}, 2020.

\bibitem{wang2021pyramid}
Wenhai Wang, Enze Xie, Xiang Li, Deng-Ping Fan, Kaitao Song, Ding Liang, Tong
  Lu, Ping Luo, and Ling Shao.
\newblock Pyramid vision transformer: A versatile backbone for dense prediction
  without convolutions.
\newblock {\em arXiv}, 2021.

\bibitem{wang2018non}
Xiaolong Wang, Ross Girshick, Abhinav Gupta, and Kaiming He.
\newblock Non-local neural networks.
\newblock In {\em CVPR}, 2018.

\bibitem{wu2019pay}
Felix Wu, Angela Fan, Alexei Baevski, Yann~N Dauphin, and Michael Auli.
\newblock Pay less attention with lightweight and dynamic convolutions.
\newblock In {\em ICLR}, 2019.

\bibitem{xie2017aggregated}
Saining Xie, Ross Girshick, Piotr Doll{\'a}r, Zhuowen Tu, and Kaiming He.
\newblock Aggregated residual transformations for deep neural networks.
\newblock In {\em CVPR}, 2017.

\bibitem{Xu2021CoScaleCI}
Weijian Xu, Yifan Xu, Tyler Chang, and Zhuowen Tu.
\newblock Co-scale conv-attentional image transformers.
\newblock {\em arXiv}, 2021.

\bibitem{yang2020transpose}
Sen Yang, Zhibin Quan, Mu Nie, and Wankou Yang.
\newblock Transpose: Towards explainable human pose estimation by transformer.
\newblock {\em arXiv}, 2020.

\bibitem{yuan2021incorporating}
Kun Yuan, Shaopeng Guo, Ziwei Liu, Aojun Zhou, Fengwei Yu, and Wei Wu.
\newblock Incorporating convolution designs into visual transformers.
\newblock {\em arXiv}, 2021.

\bibitem{yuan2019object}
Yuhui Yuan, Xilin Chen, and Jingdong Wang.
\newblock Object-contextual representations for semantic segmentation.
\newblock In {\em ECCV}, 2020.

\bibitem{zaheer2020big}
Manzil Zaheer, Guru Guruganesh, Avinava Dubey, Joshua Ainslie, Chris Alberti,
  Santiago Ontanon, Philip Pham, Anirudh Ravula, Qifan Wang, Li Yang, et~al.
\newblock Big bird: Transformers for longer sequences.
\newblock In {\em NeurIPS}, 2020.

\bibitem{Zhang2021MultiScaleVL}
Pengchuan Zhang, Xiyang Dai, Jianwei Yang, Bin Xiao, Lu Yuan, Lei Zhang, and
  Jianfeng Gao.
\newblock Multi-scale vision longformer: A new vision transformer for
  high-resolution image encoding.
\newblock {\em arXiv}, 2021.

\bibitem{zhang2021rest}
Qinglong Zhang and Yubin Yang.
\newblock Rest: An efficient transformer for visual recognition.
\newblock {\em arXiv preprint arXiv:2105.13677}, 2021.

\bibitem{zhao2019muse}
Guangxiang Zhao, Xu Sun, Jingjing Xu, Zhiyuan Zhang, and Liangchen Luo.
\newblock Muse: Parallel multi-scale attention for sequence to sequence
  learning.
\newblock {\em arXiv preprint arXiv:1911.09483}, 2019.

\bibitem{zhao2021proto}
Zelin Zhao, Karan Samel, Binghong Chen, and Le Song.
\newblock Proto: Program-guided transformer for program-guided tasks.
\newblock {\em arXiv preprint arXiv:2110.00804}, 2021.

\bibitem{zheng2020end}
Minghang Zheng, Peng Gao, Xiaogang Wang, Hongsheng Li, and Hao Dong.
\newblock End-to-end object detection with adaptive clustering transformer.
\newblock {\em arXiv}, 2020.

\bibitem{zheng2020rethinking}
Sixiao Zheng, Jiachen Lu, Hengshuang Zhao, Xiatian Zhu, Zekun Luo, Yabiao Wang,
  Yanwei Fu, Jianfeng Feng, Tao Xiang, Philip H.~S. Torr, and Li Zhang.
\newblock Rethinking semantic segmentation from a sequence-to-sequence
  perspective with transformers.
\newblock In {\em CVPR}, 2021.

\bibitem{zhu2020deformable}
Xizhou Zhu, Weijie Su, Lewei Lu, Bin Li, Xiaogang Wang, and Jifeng Dai.
\newblock Deformable detr: Deformable transformers for end-to-end object
  detection.
\newblock In {\em ICLR}, 2021.

\bibitem{zhu2019asymmetric}
Zhen Zhu, Mengde Xu, Song Bai, Tengteng Huang, and Xiang Bai.
\newblock Asymmetric non-local neural networks for semantic segmentation.
\newblock In {\em ICCV}, 2019.

\end{thebibliography}
}
\end{document}